\documentclass[11pt]{article}

\usepackage[final]{acl}

\usepackage{times}
\usepackage{latexsym}

\usepackage[T1]{fontenc}

\usepackage[utf8]{inputenc}

\usepackage{microtype}

\usepackage{inconsolata}

\usepackage{graphicx}

\usepackage{amsfonts}
\usepackage{amsmath}
\usepackage{booktabs}
\usepackage{enumitem}
\usepackage{makecell}
\usepackage{multirow}
\usepackage{subcaption}
\usepackage{xurl}

\newcommand{\ours}{SpecVocab}
\newcommand{\result}[2]{#1\textsubscript{\,\textcolor{gray}{#2}}}

\title{Speculative Decoding with a Speculative Vocabulary}

\author{
\textbf{Miles Williams}\normalfont{\textsuperscript{1,2}}\quad
\textbf{Young D. Kwon}\normalfont{\textsuperscript{2}}\quad
\textbf{Rui Li}\normalfont{\textsuperscript{2}} \\
\textbf{Alexandros Kouris}\normalfont{\textsuperscript{2}}\quad
\textbf{Stylianos I. Venieris}\normalfont{\textsuperscript{2}} \\ \\
\textsuperscript{1}University of Sheffield\quad
\textsuperscript{2}Samsung AI Center, Cambridge, UK \\
\small{\textbf{Correspondence:} \href{mailto:mil.williams@samsung.com}{mil.williams@samsung.com}}
}

\begin{document}
\maketitle
\begin{abstract}
Speculative decoding has rapidly emerged as a leading approach for accelerating language model (LM) inference, as it offers substantial speedups while yielding identical outputs. This relies upon a small draft model, tasked with predicting the outputs of the target model. State-of-the-art speculative decoding methods use a draft model comprising a single decoder layer and output embedding matrix, with the latter dominating drafting time for the latest LMs. Recent work has sought to address this output distribution bottleneck by reducing the vocabulary of the draft model. While this can improve throughput, it compromises speculation effectiveness when the target token is out-of-vocabulary. In this paper, we argue for vocabulary speculation as an alternative to a reduced vocabulary. We propose \ours{}, an efficient and effective method that selects a vocabulary subset per decoding step. Across a variety of tasks, we show that \ours{} can achieve a higher acceptance length than state-of-the-art speculative decoding method, EAGLE-3. Notably, this yields up to an 8.1\% increase in average throughput over EAGLE-3.\footnote{\url{https://github.com/SamsungLabs/SpecVocab}}
\end{abstract}

\section{Introduction}

Despite the remarkable capabilities of large language models (LMs) \citep{kamath-etal-2025-gemma, yang-etal-2025-qwen3, agarwal-etal-2025-gpt}, their autoregressive design continues to limit inference efficiency. Speculative decoding has emerged as a prominent approach to accelerate inference while maintaining identical outputs \citep{xia-etal-2024-unlocking}. Conventionally, speculative decoding combines the desired \emph{target} model with a smaller \emph{draft} model. The draft model rapidly generates a series of candidate tokens that are then verified in parallel via a single forward pass of the target model \citep{leviathan-etal-2023-fast, chen-etal-2023-accelerating-large}. %

Contemporary speculative decoding methods leverage lightweight draft models for efficient drafting \citep{miao-etal-2024-specinfer, cheng-etal-2024-recurrent, wertheimer-etal-2024-accelerating, li-etal-2024-eagle1, li-etal-2024-eagle, li-etal-2025-eagle3, zhang-etal-2025-learning-harmonized}. However, \citet{zhao-etal-2025-fr} recently identified that in the widely adopted EAGLE speculative decoding framework \citep{li-etal-2024-eagle}, the majority of drafting time is spent computing the output distribution over the target vocabulary. This presents a substantial problem, as LMs continue to require expansive vocabularies \citep{tao-etal-2024-scaling, huang-etal-2025-over, takase-etal-2025-large}.

To mitigate the vocabulary bottleneck during drafting, recent work has sought to exploit the Zipfian distribution of natural language \citep{zipf-1949-human}. In theory, rare tokens are less likely to be predicted by the target model, so they can be excluded from the draft model vocabulary. FR-Spec \citep{zhao-etal-2025-fr}, EAGLE-3 \citep{li-etal-2025-eagle3}, and VocabTrim \citep{goel-etal-2025-vocabtrim} all use a fixed subset of the target model vocabulary to reduce the latency of the vocabulary projection. Nonetheless, when the next token falls outside of this subset, the current and subsequent draft tokens will be rejected, thereby eliminating the speedup from speculative decoding.

\begin{figure}[t]
\centering
\includegraphics[width=0.925\linewidth]{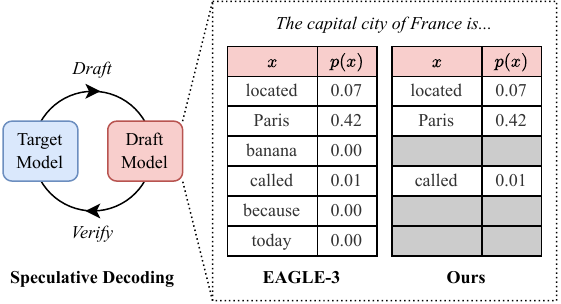}
\caption{\emph{Vocabulary speculation} accelerates speculative decoding by computing the output distribution for only a contextually relevant subset of the vocabulary.}
\label{fig:diagram}
\end{figure}

In this paper, we argue that speculative decoding should speculate not only on the next tokens, \emph{but also on the output vocabulary} (Figure~\ref{fig:diagram}). In contrast to earlier approaches, this allows the cost of computing the output distribution to be reduced, while better preserving the quantity of accepted draft tokens. Our core contributions are as follows:
\begin{enumerate}
\item We propose \ours{}, an efficient method to predict a subset of the vocabulary that is contextually relevant to the next token.

\item Across a variety of tasks, \ours{} achieves higher acceptance lengths than static vocabulary methods (\textit{i.e.}~EAGLE-3, FR-Spec, and VocabTrim). This enables up to 8.1\% higher average throughput than EAGLE-3.

\item We implement and benchmark a custom kernel that accelerates logits computation for vocabulary subsets by up to 5$\times$.

\item We empirically demonstrate that the reduced-vocabulary draft model training process introduced in EAGLE-3 \citep{li-etal-2025-eagle3} can unnecessarily harm draft model performance.
\end{enumerate}

\begin{figure*}[t]
\centering
\includegraphics[width=\linewidth]{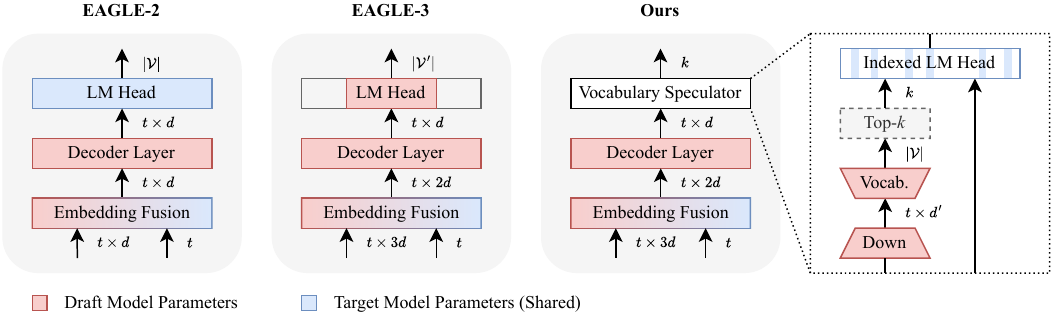}
\caption{Outline of the draft model architectures for speculative decoding. EAGLE-2 forms predictions over the entire target model vocabulary, whereas EAGLE-3 uses a fixed subset, as in FR-Spec and VocabTrim. In contrast, \ours{} (ours) speculates on which subset of the target model vocabulary to use at each decoding step.}
\label{fig:architecture}
\end{figure*}

\section{Related Work}

\paragraph{Vocabulary representations.}

Recent LMs leverage increasingly large subword vocabularies. Early Transformer-based LMs such as BERT \citep{devlin-etal-2019-bert} and GPT-2 \citep{radford-etal-2019-language} used vocabularies with 30K and 50K tokens, respectively. In comparison, more recent models such as OLMo 2 \citep{walsh-etal-2025-2-olmo} and Qwen3 \citep{yang-etal-2025-qwen3} adopt larger vocabularies of 100K and 152K, respectively. Recent work has highlighted the performance benefits of pre-training LMs with such substantial vocabularies \citep{tao-etal-2024-scaling, huang-etal-2025-over, takase-etal-2025-large}.

\paragraph{Speculative decoding.}

The core mechanism behind speculative decoding is to propose multiple tokens and continue generation from the longest correct prefix \citep{stern-etal-2018-blockwise, sun-etal-2021-instantaneous}. This has been popularized as the \emph{draft-then-verify} pattern, using an efficient draft model, while preserving the output distribution of the target model \citep{xia-etal-2023-speculative, leviathan-etal-2023-fast, chen-etal-2023-accelerating-large}. Recent speculative decoding approaches have explored using auxiliary heads \citep{cai-etal-2024-medusa, ankner-etal-2024-hydra}, contextual embeddings \citep{gritta-etal-2025-dresd}, intermediate hidden states \citep{cheng-etal-2024-recurrent, li-etal-2025-eagle3}, and tree-like drafting \citep{spector-re-2023-accelerating, li-etal-2025-eagle3}. In particular, the EAGLE series of speculative decoding methods \citep{li-etal-2024-eagle1, li-etal-2024-eagle, li-etal-2025-eagle3} have been widely adopted, both in popular inference engines \citep{kwon-etal-2023-efficient, zheng-etal-2024-sglang} and at scale \citep{tang-etal-2025-efficient}.

\paragraph{Reduced-vocabulary draft models.}

While speculative decoding methods have sought to maximize the efficiency of the draft models, the overhead from computing the output distribution over the model vocabulary has persisted. \citet{zhao-etal-2025-fr} first identified the output distribution bottleneck in speculative decoding draft models. They proposed FR-Spec, which prunes the embeddings for less common tokens, exploiting the long-tailed nature of natural language frequency distributions \citep{zipf-1949-human}. Independently, \citet{goel-etal-2025-vocabtrim} proposed VocabTrim, also pruning the output embedding matrix based on token frequency. In contrast to FR-Spec, which suggests using large-scale pre-training corpora \citep{soboleva-etal-2023-slimpajama} for token frequency computation, VocabTrim leverages synthetic data generated by the target model. Finally, EAGLE-3 \citep{li-etal-2025-eagle3} also adopts a reduced vocabulary based on synthetic data from the target model. However, the output embedding matrix is trained from scratch, rather than being pruned after training. \citet{zhao-etal-2025-fr} find that a vocabulary size of 32K provides the optimal throughput, with this vocabulary size also being used by EAGLE-3. However, a fixed vocabulary subset is inherently context-agnostic, leading to suboptimal performance when output tokens fall outside the subset. Our work addresses this through context-aware vocabulary speculation, selecting a relevant subset at each decoding step.

\section{Vocabulary Speculation}

\subsection{Preliminaries}

Speculative decoding accelerates autoregressive generation by introducing a lightweight draft model $q$ that generates candidate sequences, which are subsequently verified by the target model $p$. The draft model has an output embedding matrix (\textit{i.e.}~LM head) $\mathbf{U} \in \mathbb{R}^{|\mathcal{V}|\times d}$, where $|\mathcal{V}|$ is the vocabulary size of the target model and $d$ is the draft model dimensionality. At each decoding step $t$, the draft model produces logits $\mathbf{z}_t = \mathbf{U}\mathbf{h}_t$, where $\mathbf{h}_t$ is the final hidden state of the draft model. These logits are then used to form the next-token probability distribution $q(\mathbf{x}_t \mid \mathbf{x}_{<t}) = \text{softmax}(\mathbf{z}_t) \in \mathbb{R}^{|\mathcal{V}|}$.

\subsection{Problem Definition}

Computing the output distribution proves to be an expensive operation, as it scales with both the vocabulary size $|\mathcal{V}|$ and model dimensionality $d$ \citep{zhao-etal-2025-fr}. \emph{Vocabulary speculation} seeks to reduce this cost by efficiently predicting a contextually relevant subset of the target model vocabulary. We formally define the vocabulary speculation problem as follows:
\begin{quote}
\emph{Given a context $c$, which subset of the vocabulary $\mathcal{K} \subset \mathcal{V}$ should be evaluated to support accurate decoding?}
\end{quote}
The subset $\mathcal{K}$ should be compact, with $|\mathcal{K}| \ll |\mathcal{V}|$, while providing sufficient coverage of the tokens likely to be selected by the sampler for $p(\mathbf{x}_t \mid \mathbf{x}_{<t})$.

\subsection{\ours{}}

We propose \ours{}, a method for vocabulary speculation that incorporates an efficient vocabulary ranking module to predict the most relevant vocabulary subset (Figure~\ref{fig:architecture}). 

\paragraph{Step 1.}

We start by computing an approximate ranking over the target vocabulary to inform candidate selection. We elect to use the final hidden state from the draft model $\mathbf{h}_t$ as our context. First, we obtain a reduced-dimensionality intermediate hidden state $\mathbf{h}'_t$ where $\mathbf{W}_\text{down} \in \mathbb{R}^{d' \times d}$ and $d' \ll d$. This is used to efficiently compute approximate logits $\mathbf{s}_t$ over the vocabulary, where $\mathbf{W}_\text{vocab} \in \mathbb{R}^{|\mathcal{V}| \times d'}$:
\begin{align*}
\mathbf{h}'_t &= \mathbf{W}_\text{down}\mathbf{h}_t \\
\mathbf{s}_t &= \mathbf{W}_\text{vocab}\mathbf{h}'_t
\end{align*}

\paragraph{Step 2.}

We then select the top-$k$ indices from $\mathbf{s}_t$, forming our candidate vocabulary $\mathcal{K}_t$, where $\mathbf{k}_t$ represents the indices of the corresponding tokens. For simplicity, we use a fixed value of $k$, although this value can also be varied per decoding step.
\begin{align*}
\mathbf{k}_t &= \text{top-k}(\mathbf{s}_t, k)
\end{align*}
As \ours{} is context-aware, it can use substantially smaller subsets than prior work (typically $k = 2048$), compared with the 32K subset conventionally used in static vocabulary methods~\citep{zhao-etal-2025-fr,goel-etal-2025-vocabtrim}.

\paragraph{Step 3.}

Finally, we compute the output logits $\mathbf{z}'_t$ for the candidate vocabulary using the original hidden states $\mathbf{h}_t$. Here, $\mathbf{U}'_t$ represents an intermediate matrix containing only the output embeddings corresponding to the candidate vocabulary.
\begin{align*}
\mathbf{U}'_t &= \mathbf{U}[\mathbf{k}_t, :] \\
\mathbf{z}'_t &= \mathbf{U}'_t\mathbf{h}_t
\end{align*}
Therefore, the final output distribution produced by the draft model over the candidate vocabulary is:
\begin{equation*}
q_{\mathcal{K}_t}(\mathbf{x}_t \mid \mathbf{x}_{<t}) = \text{softmax}(\mathbf{z}'_t)
\end{equation*}

\subsection{Training}

To enable accurate vocabulary speculation, we learn the weights for $\mathbf{W}_\text{down}$ and $\mathbf{W}_\text{vocab}$ through distillation from the target model. During training, we exclude \textit{Steps 2 \& 3}, focusing only on approximating the target model output distribution. Specifically, we form the output distribution over the entire target model vocabulary:
\begin{equation*}
q_\text{aux}(\mathbf{x}_t \mid \mathbf{x}_{<t}) = \text{softmax}(\mathbf{s}_t)
\end{equation*}
The vocabulary speculation module is then trained alongside the draft model. We formulate the final loss $\mathcal{L}$ as a combination of the draft model and vocabulary speculator losses, where the contribution of the latter is modulated by the weight $\lambda$. The intuition behind this approach is that we wish to encourage the draft model to learn somewhat compressed representations for the final hidden state, without compromising overall acceptance length.
\begin{equation*}
\mathcal{L} = \mathcal{L}_\text{TTT}(p, q) + \lambda \mathcal{L}_\text{TTT}(p, q_\text{aux})
\end{equation*}
We use $\mathcal{L}_\text{TTT}$ to denote the \emph{training-time test} loss proposed in EAGLE-3 \citep{li-etal-2025-eagle3}, which can be conceptualized as a cross-entropy loss across additional timesteps simulated during training. However, we emphasize that our method is not tied to any specific training approach. Rather, we simply mirror the same loss used for the draft model to the vocabulary speculator module.

\subsection{Inference}

To demonstrate the efficacy of our approach, we integrate \ours{} with a state-of-the-art speculative decoding method, EAGLE-3 \citep{li-etal-2025-eagle3}. EAGLE-3 leverages tree attention to simultaneously predict multiple tokens at each decoding step of the draft model. Consequently, in practice, \ours{} operates on batches of hidden states following the tree-like draft. Nonetheless, we emphasize that our approach is not tied to EAGLE-3 and can be applied to other speculative decoding approaches that leverage lightweight draft models.

\paragraph{Custom kernel.}

\begin{figure}[t]
\centering
\begin{subfigure}{\linewidth}
\centering
\includegraphics[width=\linewidth]{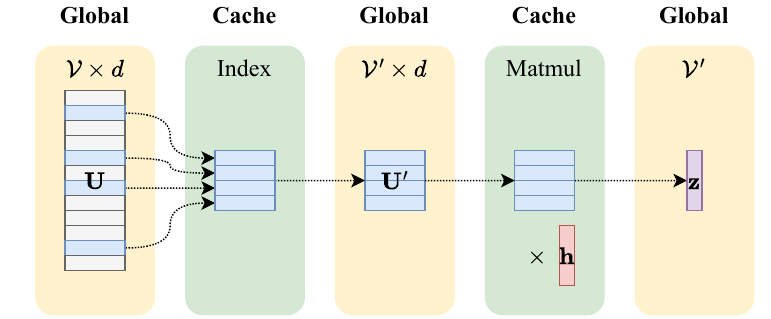}
\caption{Indexed LM head operation in PyTorch.}
\end{subfigure}
\begin{subfigure}{\linewidth}
\centering
\includegraphics[width=\linewidth]{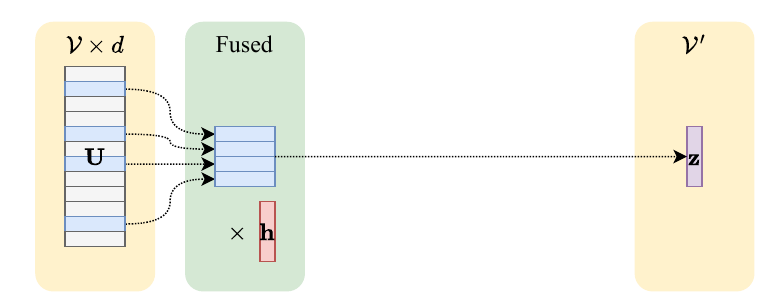}
\caption{Our custom fused kernel implementation.}
\end{subfigure}
\caption{The sequence of memory access operations required by the indexed LM head operation, alternating between global and cache memory.}
\label{fig:kernel-diagram}
\end{figure}

In \emph{Step 3}, computing the output logits requires taking the dot product between the final hidden state $\mathbf{h}_t$ and the embedding at each index of $\mathbf{k}_t$ in $\mathbf{U}$. Naively, this can be performed by materializing the dense matrix $\mathbf{U}'_t$ followed by a matrix-vector multiplication with $\mathbf{h}_t$. However, this would require reading, writing, and re-reading $k$ vectors from memory, along with any allocation costs. To avoid this unnecessary memory traffic, we instead implement a fused kernel that only reads each embedding from memory once. We visually illustrate these memory operations and the difference between the two approaches in Figure~\ref{fig:kernel-diagram}.

\section{Experimental Setup}

\paragraph{Baselines.}

To evaluate the performance of our approach, we compare against a variety of baselines:

\begin{itemize}[left=0pt]
\item \textbf{Autoregressive Decoding}. To contextualize the improvement in throughput from speculative decoding, we report results for standard autoregressive decoding with the target model.

\item \textbf{EAGLE-3} \citep{li-etal-2025-eagle3}. We focus on the state-of-the-art EAGLE-3 method for speculative decoding. For transparency, we report results for both our own reproduction of draft model training and a third-party model.\footnote{\url{https://huggingface.co/Tengyunw/qwen3_8b_eagle3}}

\item \textbf{FR-Spec} \citep{zhao-etal-2025-fr}. We perform post-training vocabulary pruning with FR-Spec. Following the original work, we compute token frequencies with a 1B-token sample of the SlimPajama pre-training corpus \citep{soboleva-etal-2023-slimpajama}.

\item \textbf{VocabTrim} \citep{goel-etal-2025-vocabtrim}. Similarly, we also perform post-training vocabulary pruning with VocabTrim. Following the original work, we compute the token frequencies with target model data. Therefore, this uses the same synthetic dataset used to train all of our draft models.
\end{itemize}

\noindent We emphasize that both FR-Spec and VocabTrim build upon EAGLE-2 \citep{li-etal-2024-eagle}, rather than EAGLE-3. For a fair comparison, we build stronger baselines by re-implementing both methods with EAGLE-3. Namely, we apply each pruning method once to a full-vocabulary EAGLE-3 draft model. Following \citet{zhao-etal-2025-fr} and \citet{li-etal-2025-eagle3}, we use a draft model vocabulary size of 32K.

\begin{table*}[t]
\small
\centering
\begin{tabular}{l|rrrrrr|r}
\toprule
Method & \multicolumn{1}{c}{Conv.} & \multicolumn{1}{c}{MT} & \multicolumn{1}{c}{Summ.} & \multicolumn{1}{c}{QA} & \multicolumn{1}{c}{Math} & \multicolumn{1}{c}{RAG} & \multicolumn{1}{|c}{Mean} \\
\midrule
Autoregressive Decoding & 1.00 & 1.00 & 1.00 & 1.00 & 1.00 & 1.00 & 1.00 \\
EAGLE-3 & 4.52 & 3.34 & 3.86 & 4.46 & 5.28 & 4.52 & 4.33 \\
EAGLE-3 (Our Reproduction) & 5.11 & 3.60 & 4.21 & 4.80 & 5.94 & 5.04 & 4.78 \\
\midrule
EAGLE-3 + FR-Spec & 4.62 & 3.80 & \textbf{4.52} & 4.52 & 5.77 & 4.90 & 4.69 \\
EAGLE-3 + VocabTrim & 5.17 & 3.69 & 4.28 & 4.72 & 6.04 & 5.13 & 4.84 \\
EAGLE-3 + \ours{} (Ours) & \textbf{5.30} & \textbf{3.82} & 4.49 & \textbf{4.92} & \textbf{6.28} & \textbf{5.26} & \textbf{5.01} \\
\bottomrule
\end{tabular}
\caption{Acceptance length (tokens) for each decoding method across the Spec-Bench benchmark with Qwen3 8B. We report the average acceptance length over five seeds. The best result in each category is highlighted in \textbf{bold}.}
\label{tab:acceptance-length-qwen3-8b}
\end{table*}

\begin{table*}[t]
\small
\centering
\begin{tabular}{l|rrrrrr|r}
\toprule
Method & \multicolumn{1}{c}{Conv.} & \multicolumn{1}{c}{MT} & \multicolumn{1}{c}{Summ.} & \multicolumn{1}{c}{QA} & \multicolumn{1}{c}{Math} & \multicolumn{1}{c}{RAG} & \multicolumn{1}{|c}{Mean} \\
\midrule
Autoregressive Decoding & \result{98.5}{0.0} & \result{92.4}{0.1} & \result{96.5}{0.0} & \result{98.9}{0.0} & \result{98.7}{0.0} & \result{97.3}{0.0} & \result{97.1}{0.0} (1.00$\times$) \\
EAGLE-3 & \result{227.7}{1.6} & \result{149.7}{0.4} & \result{187.1}{0.9} & \result{226.3}{0.9} & \result{264.1}{1.4} & \result{221.5}{1.5} & \result{212.7}{0.8} (2.19$\times$) \\
EAGLE-3 (Our Reproduction) & \result{258.9}{0.1} & \result{159.1}{0.2} & \result{203.9}{0.2} & \result{244.1}{0.2} & \result{297.8}{0.2} & \result{246.9}{0.2} & \result{235.1}{0.2} (2.42$\times$) \\
\midrule
EAGLE-3 + FR-Spec & \result{234.2}{0.2} & \result{166.0}{0.3} & \result{\textbf{218.2}}{0.2} & \result{230.2}{0.2} & \result{289.3}{0.3} & \result{241.3}{0.2} & \result{229.8}{0.2} (2.37$\times$) \\
EAGLE-3 + VocabTrim & \result{261.7}{0.2} & \result{162.0}{0.2} & \result{207.2}{0.1} & \result{239.7}{0.1} & \result{302.9}{0.2} & \result{252.1}{0.2} & \result{237.6}{0.2} (2.45$\times$) \\
EAGLE-3 + \ours{} (Ours) & \result{\textbf{267.6}}{0.2} & \result{\textbf{166.5}}{0.2} & \result{216.5}{0.1} & \result{\textbf{249.4}}{0.2} & \result{\textbf{313.6}}{0.2} & \result{\textbf{257.2}}{0.2} & \result{\textbf{245.2}}{0.2} (2.53$\times$) \\
\bottomrule
\end{tabular}
\caption{Throughput (tokens per second) for each decoding method across the Spec-Bench tasks with Qwen3 8B. We report the average throughput over five seeds, with the standard deviation as subscripts. The best result in each category is highlighted in \textbf{bold}.}
\label{tab:throughput-qwen3-8b}
\end{table*}

\paragraph{Models.}

To examine how each approach generalizes, we experiment with four open-source LMs from two separate model families. We employ Qwen3 \citep{yang-etal-2025-qwen3} in 4B and 8B sizes, and OLMo 2 \citep{walsh-etal-2025-2-olmo} in 1B and 7B sizes. We select these model families as they represent state-of-the-art performance at the time of writing, relative to their size and respective open-source categories (open-weights and fully-open).

\paragraph{Training.}

We closely follow the original training protocol from \citet{li-etal-2025-eagle3}. We adopt the UltraChat \citep{ding-etal-2023-enhancing} dataset, with assistant responses generated by the target model, providing approximately 464K training examples. We report the training hyperparameters in Appendix~\ref{app:hyperparameters}.

\paragraph{Evaluation.}

We adopt the Spec-Bench \citep{xia-etal-2024-unlocking} dataset in its entirety for evaluation and benchmarking. This consists of a varied set of tasks, namely multi-turn conversation (MT-Bench; \citealp{zheng-etal-2023-judging}), machine translation (WMT14 DE-EN; \citealp{bojar-etal-2014-findings}), mathematical reasoning (GSM8K; \citealp{cobbe-etal-2021-training}), summarization (CNN/Daily Mail; \citealp{hermann-etal-2015-teaching, see-etal-2017-get}), retrieval-augmented generation, and question answering (Natural Questions; \citealp{kwiatkowski-etal-2019-natural}). We report full dataset statistics in Appendix~\ref{app:datasets}.

\paragraph{Metrics.}

Following convention \citep{xia-etal-2024-unlocking}, we focus our evaluation on two key metrics: acceptance length and throughput. \emph{Acceptance length} is the average number of tokens proposed by the draft model that are successfully verified by the target model, reflecting the draft model accuracy. However, an increased acceptance length will not translate to faster decoding if the draft model execution is too slow \citep{zhao-etal-2025-fr}. Consequently, we also measure system \emph{throughput}, the rate at which output tokens are generated.

\paragraph{Implementation details.}

To ensure that our experimental results reflect real-world performance, we implement all experiments using the highly-optimized SGLang inference framework \citep{zheng-etal-2024-sglang}. Notably, SGLang leverages CUDA graphs to virtually eliminate the CPU overhead when launching GPU operations. We highlight this point as \citet{zhao-etal-2025-fr} find that the performance loss incurred by large vocabularies can be obscured by CPU overhead in naive implementations.

\paragraph{Computational resources.}

We perform all model training using four NVIDIA H100 80GB GPUs, while we perform all inference experiments using only a single GPU. For consistency, we ensure that all evaluations for a given model configuration are performed on the same underlying physical hardware.

\section{Results \& Discussion}
\label{sec:results}

\begin{figure*}
\centering
\includegraphics{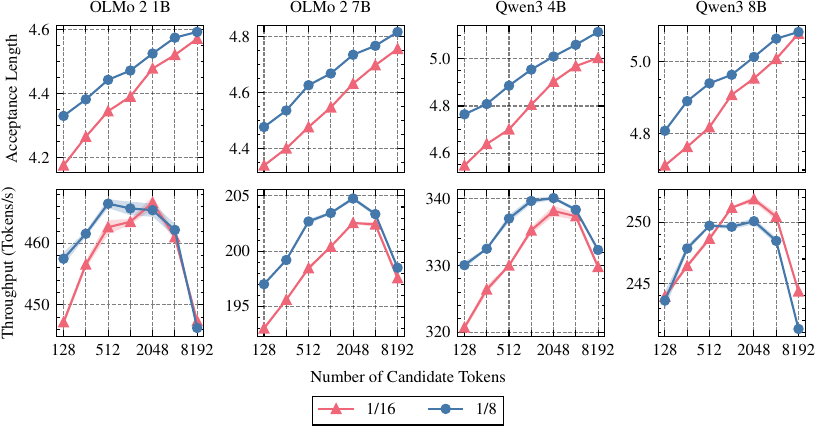}
\caption{The acceptance length and throughput when varying both the number of candidate tokens ($k$) and intermediate dimensionality of our method relative to the target model ($d'/d$). We present the results for every model across five seeds, with the standard deviation denoted by the shaded area.}
\label{fig:ablation-num-candidates}
\end{figure*}

\paragraph{\ours{} consistently outperforms \mbox{EAGLE-3}.}

Table~\ref{tab:acceptance-length-qwen3-8b} presents the acceptance length across the Spec-Bench tasks for Qwen3 8B. Across all tasks, we observe that \ours{} demonstrates a higher acceptance length than EAGLE-3. This ranges from a 2.5\% improvement for question answering (+0.12 tokens per drafting step) to a 6.8\% improvement for summarization (+0.29 tokens). Considering the 4B size of Qwen3 (Table~\ref{tab:acceptance-length-qwen3-4b}, Appendix~\ref{app:complete-results}), we observe that acceptance length ranges from a 3.1\% improvement for question answering to 10.3\% for retrieval-augmented generation. On average, \ours{} achieves 4.8\% higher acceptance length (+0.23 tokens per drafting step) for Qwen3 8B, while Qwen3 4B achieves a slightly higher 6.0\% improvement (+0.28 tokens).

\paragraph{Crucially, improvements in acceptance length translate to greater throughput for \ours{}.}

Considering all four models, we observe that there are substantial improvements in average throughput compared to EAGLE-3. Specifically, we note a 4.3\% improvement for Qwen3 8B (Table~\ref{tab:throughput-qwen3-8b}), and a 5.0\% increase for Qwen3 4B (Table~\ref{tab:throughput-qwen3-4b}, Appendix~\ref{app:complete-results}). For OLMo 2 1B, we observe a similar improvement of 5.0\%, while OLMo 2 7B achieves an even greater increase of 8.1\% (Tables \ref{tab:throughput-olmo2-1b} and \ref{tab:throughput-olmo2-7b}, Appendix~\ref{app:complete-results}). We believe that these substantial increases in throughput, measured in a real-world setting, underscore the efficacy of our method.

\paragraph{Post-training vocabulary pruning outperforms the reduced-vocabulary training of EAGLE-3.}

We examine the difference in acceptance length between EAGLE-3 and VocabTrim, which both rely on the token frequency statistics from the target-generated training data. However, EAGLE-3 performs vocabulary pruning before training, while VocabTrim is applied afterwards. We observe that VocabTrim outperforms EAGLE-3 in all but one case, with the exception being the question answering task for Qwen3 8B (Table~\ref{tab:acceptance-length-qwen3-8b}). Considering all models, we observe that VocabTrim achieves an average increase in acceptance length ranging from 1.1\% for Qwen3 8B (Table~\ref{tab:acceptance-length-qwen3-8b}) to 2.7\% for OLMo 2 7B (Table~\ref{tab:acceptance-length-olmo2-7b}, Appendix~\ref{app:complete-results}). \emph{We highlight the significance of this finding, which suggests that the reduced-vocabulary training regime from EAGLE-3 can unnecessarily harm draft model performance.}

\paragraph{For a static vocabulary, target model token frequencies perform better than an external corpus.}

Since FR-Spec \citep{zhao-etal-2025-fr} and VocabTrim \citep{goel-etal-2025-vocabtrim} are concurrent studies, neither offers a direct comparison. We therefore examine the performance difference between token frequencies computed from large-scale corpora versus data generated by the target model, respectively. Considering the resulting throughput across all models, we observe that there are clear trends across tasks (Tables \ref{tab:throughput-qwen3-8b}, \ref{tab:throughput-olmo2-1b}, \ref{tab:throughput-olmo2-7b}, and \ref{tab:throughput-qwen3-4b}). Specifically, VocabTrim performs better for multi-turn conversation, question answering, mathematical reasoning, and retrieval-augmented generation, whereas FR-Spec performs better for machine translation and summarization. As VocabTrim performs better in the majority of tasks (i.e. four of six), this corroborates the finding from \citet{goel-etal-2025-vocabtrim} that using target-generated token frequencies benefits performance. 

\paragraph{\ours{} generally outperforms all methods.}

We first consider the throughput for each of the Spec-Bench tasks individually, across all models. We observe that \ours{} outperforms all other methods in all but one case, i.e. 23 of 24 instances. The only exception is the summarization task for Qwen3 8B (Table~\ref{tab:throughput-qwen3-8b}), where \ours{} is narrowly slower than FR-Spec by 0.8\% (1.7 fewer tokens per second). Considering the average throughput across tasks, we observe that \ours{} consistently outperforms all other methods across every model. Next, we consider the performance of our approach over VocabTrim, the best-performing static vocabulary method. We observe that the improvements in average throughput for \ours{} range from 3.1\% for Qwen3 4B (Table~\ref{tab:throughput-qwen3-4b}, Appendix~\ref{app:complete-results}) to 5.4\% for OLMo 2 7B (Table~\ref{tab:throughput-olmo2-7b}, Appendix~\ref{app:complete-results}). For three of the four models, this translates to at least an additional ten tokens per second.

\paragraph{Performance improvements vary between tasks.}

Across all models, we observe that the improvements in throughput differ between tasks. For example, we first consider the range of improvements in throughput demonstrated by our method versus EAGLE-3, across all models. The greatest improvement is 11.0\% on mathematical reasoning with OLMo 2 7B (Table~\ref{tab:throughput-olmo2-7b}, Appendix~\ref{app:complete-results}). In contrast, the lowest improvement is 2.2\% on question answering with Qwen3 8B (Table~\ref{tab:throughput-qwen3-8b}). This pattern is not consistent across models. For example, the greatest improvement in throughput for OLMo 2 1B is 8.0\% on the summarization task (Table~\ref{tab:throughput-olmo2-1b}, Appendix~\ref{app:complete-results}), while Qwen3 4B sees a maximum improvement of 9.5\% for the retrieval-augmented generation task (Table~\ref{tab:throughput-qwen3-4b}, Appendix~\ref{app:complete-results}).

\paragraph{\ours{} performance scales with model size.}

To examine how model size impacts the performance of \ours{}, we compare the throughput speedup between models of different sizes from the same model family. Following the trend set by EAGLE-3, we observe that our method can achieve a greater speedup for larger models. For example, \ours{} achieves a 2.18$\times$ speedup over autoregressive decoding for Qwen3 4B (Table~\ref{tab:throughput-qwen3-4b}, Appendix~\ref{app:complete-results}), yet achieves a larger speedup of 2.53$\times$ for Qwen3 8B (Table~\ref{tab:throughput-qwen3-8b}). In the case of OLMo 2 1B, we observe a 1.38$\times$ speedup for \ours{} over autoregressive decoding (Table~\ref{tab:throughput-olmo2-1b}, Appendix~\ref{app:complete-results}), whereas we see a much larger speedup of 2.20$\times$ for OLMo 2 7B (Table~\ref{tab:throughput-olmo2-7b}, Appendix~\ref{app:complete-results}).

\section{Analysis}
\label{sec:analysis}

\paragraph{\ours{} requires no more than 2\% of the exact logits to be computed at each decoding step.}

Figure~\ref{fig:ablation-num-candidates} presents the throughput when varying the number of candidate tokens ($k$). Across all models, we observe that computing the exact logits for only 2048 candidate tokens is sufficient for maximum throughput. This represents 1.4\% and 2.0\% of the target model vocabulary for Qwen3 and OLMo 2, respectively. In comparison to the static vocabulary methods (EAGLE-3, FR-Spec, and VocabTrim), this requires over 15$\times$ fewer tokens to be evaluated. Interestingly, for the smallest model (OLMo 2 1B), maximum throughput can be achieved by using only 512 tokens. This is equivalent to 0.5\% of the model vocabulary, over 60$\times$ fewer than the 32K subset used by the static vocabulary methods.

\paragraph{The intermediate dimensionality for \ours{} can be as small as just 1/16 of the target model.}

Figure~\ref{fig:ablation-num-candidates} also shows the acceptance length when varying the intermediate embedding dimensionality ($d'$). We observe that increasing the intermediate dimensionality consistently leads to an increase in acceptance length across all models and quantities of candidate tokens. However, this also increases the computational complexity of forming the candidate vocabulary. Therefore, increasing the dimensionality may not lead to an increase in throughput. As an example, we consider two models with identical intermediate dimensionality, OLMo 2 7B and Qwen3 8B. The difference in acceptance length between an intermediate dimensionality of 1/16 and 1/8 for OLMo 2 7B is 2.2\%, assuming 2048 candidate tokens. For Qwen3 8B, the difference is nearly halved, at 1.2\%. Consequently, OLMo 2 7B achieves greater throughput at 1/8, while Qwen3 can achieve maximum throughput at 1/16.

\begin{table}[t]
\small
\centering
\begin{tabular}{ccc}
\toprule
Dimensionality ($d'/d$) & $\lambda$ & Acceptance Length \\
\midrule
\multirow{4}{*}{1/16} & 0.01 & 4.57 \\
& 0.1 & \textbf{4.61} \\
& 0.2 & \textbf{4.61} \\
& 0.5 & 4.60 \\
\midrule
\multirow{4}{*}{1/8} & 0.01 & 4.65 \\
& 0.1 & \textbf{4.74} \\
& 0.2 & 4.71 \\
& 0.5 & 4.71 \\
\bottomrule
\end{tabular}
\caption{The impact of the training loss weighting ($\lambda$) for \ours{} upon acceptance length. We list results for both intermediate dimensionalities with Qwen3 8B.}
\label{tab:ablation-loss}
\end{table}

\paragraph{\ours{} benefits from the joint training loss.}

Table~\ref{tab:ablation-loss} presents the impact of the loss weight hyperparameter ($\lambda$) upon the draft model acceptance length during early experimentation. We observe that increasing the contribution of the loss from the vocabulary speculator module leads to a greater acceptance length. For example, raising $\lambda$ from 0.01 to 0.1 leads to a 0.09 increase in acceptance length for Qwen3 8B when using 1/8 of the target model dimensionality. We hypothesize that allowing the vocabulary speculator module loss to affect the entire draft model has a regularizing effect, encouraging the draft model to learn more compressible representations. In turn, this may improve vocabulary speculation at reduced dimensionalities.

\paragraph{\ours{} is relatively robust to the loss weight.}

We trial values of $\lambda \in \{0.01, 0.1, 0.2, 0.5\}$ for two different intermediate embedding dimensionalities. We observe that training with $\lambda = 0.1$ achieves the highest acceptance length for both dimensionalities. For 1/16 dimensionality, an increased $\lambda$ of 0.2 also achieves the highest acceptance length. In theory, using a high loss weight should harm performance, as it is a lower fidelity approximation of the true loss. Interestingly, we observe that even using a loss weight of 0.5 (i.e. contributing a third of the overall loss) is less harmful than using a very low loss weight of 0.01. For example, compared to the best-performing loss weight (0.1) for 1/8 dimensionality, $\lambda=0.5$ leads to a 0.03 decrease in acceptance length, while $\lambda=0.01$ leads to a 3$\times$ greater decrease of 0.09.

\begin{table}[t]
\small
\centering
\begin{tabular}{lc}
\toprule
Approach & Complexity \\
\midrule
Full Vocabulary (e.g. EAGLE-2) & $\mathcal{O}(|\mathcal{V}| \cdot d)$ \\
Reduced Vocabulary (e.g. FR-Spec) & $\mathcal{O}(|\mathcal{V}'| \cdot d)$ \\
\ours{} (Ours) & $\mathcal{O}(|\mathcal{V}| \cdot d' + k \cdot d)$ \\
\bottomrule
\end{tabular}
\caption{Computational complexity of calculating the output logits for each approach.}
\label{tab:computational-complexity}
\end{table}

\paragraph{Theoretical analysis.}

Table~\ref{tab:computational-complexity} presents the computational complexity for each type of approach, specifically full vocabulary (EAGLE-2), reduced vocabulary (EAGLE-3, FR-Spec, VocabTrim), and our own method. This formulation highlights the two-phase nature of our method, consisting of a reduced-dimensionality approximation ($\mathcal{O}(|\mathcal{V}| \cdot d')$) and an exact computation over a likely subset ($\mathcal{O}(k \cdot d)$). \ours{} is asymptotically more efficient than the other approaches when $|\mathcal{V}| \cdot \frac{d'}{d} + k < |\mathcal{V}'|$. In other words, our approach necessitates both (1) informative contextual representations when $d' \ll d$, and (2) sufficiently high recall that enables a small candidate set size $k$. In contrast to reduced-vocabulary methods, which require a large $|\mathcal{V}'|$ to maintain performance, \ours{} can achieve coverage of the full vocabulary at a lower cost.

\paragraph{Our kernel achieves a 3$\times$ to 5$\times$ speedup in comparison to PyTorch.}

\begin{figure}
\centering
\includegraphics{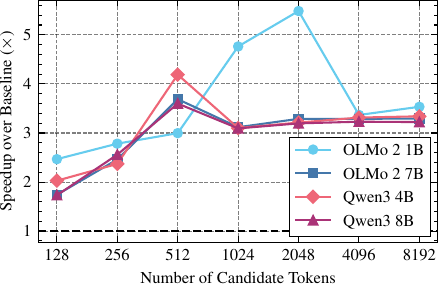}
\caption{Microbenchmark results for our custom fused kernel versus a PyTorch baseline, for each model.}
\label{fig:kernel-performance}
\end{figure}

Figure~\ref{fig:kernel-performance} presents the speedup of our kernel relative to a PyTorch baseline. To create a realistic microbenchmark, we construct input activation and weight matrices corresponding to the dimensions of each model. To eliminate the impact of CPU overhead from launching GPU operations, we perform all benchmarking using CUDA graphs. We observe that our kernel substantially outperforms the PyTorch baseline across all models and quantities of candidate tokens. For example, considering the optimal number of candidate tokens of $k=2048$, we observe that all models see a performance increase of at least 3.2$\times$. OLMo 2 1B appears to be an outlier, achieving a speedup of over 5$\times$, seemingly due to its smaller size.

\section{Conclusion}

In this paper, we argued that speculative decoding should additionally speculate on the output vocabulary for maximum performance. To this end, we proposed \ours{} as a concrete implementation of a dynamic vocabulary strategy. We empirically validated our method across a variety of tasks and models, demonstrating substantial gains in acceptance length over strong speculative decoding baselines. Crucially, these gains translated to substantial improvements in real-world throughput within a widely-used inference framework. We hope that this study will inspire further work that considers the role of vocabulary within speculative decoding.

\section*{Limitations}

\paragraph{Linguistic diversity.}

Our evaluation relies on Spec-Bench \citep{xia-etal-2024-unlocking}, which primarily targets English-language tasks. While one of the six Spec-Bench tasks involves a non-English language (German), the benchmark remains limited to West Germanic languages. As future work, we are interested in investigating how both static and dynamic vocabulary speculative decoding approaches generalize beyond the Indo-European family, particularly to languages with substantially different morphologies.

\section*{Acknowledgments}

We would like to thank the SGLang team for their optimized implementation of speculative decoding \citep{zheng-etal-2024-sglang} and the SpecForge \citep{li-etal-2025-specforge} draft model training framework. We are also grateful to the anonymous reviewers for their invaluable feedback. MW is supported by the Centre for Doctoral Training in Speech and Language Technologies (SLT) and their Applications funded by UK Research and Innovation grant EP/S023062/1.

\bibliography{custom, anthology-1, anthology-2}

\appendix

\section{Complete Results}
\label{app:complete-results}

In addition to the results presented for Qwen3 8B in Section~\ref{sec:results} (Tables \ref{tab:acceptance-length-qwen3-8b} and \ref{tab:throughput-qwen3-8b}), we report complete results for the remaining models. Tables \ref{tab:acceptance-length-olmo2-1b}, \ref{tab:acceptance-length-olmo2-7b}, and \ref{tab:acceptance-length-qwen3-4b} present the acceptance length with each decoding method across the Spec-Bench benchmark for OLMo 2 1B, OLMo 2 7B, and Qwen3 4B, respectively. Tables \ref{tab:throughput-olmo2-1b}, \ref{tab:throughput-olmo2-7b}, and \ref{tab:throughput-qwen3-4b} report the corresponding throughput values. We provide a discussion of these results in Section~\ref{sec:results}.

\section{Additional Experiments}

\subsection{EAGLE-3 with Full Vocabulary}

To establish an upper bound on acceptance length, we construct an additional EAGLE-3 baseline that adopts the complete vocabulary of the target model. Specifically, we follow the implementation from EAGLE-2 \citep{li-etal-2024-eagle} in sharing the target model LM head with the draft model (Figure~\ref{fig:architecture}). Tables~\ref{tab:acceptance-length-qwen3-8b-full-vocabulary} and \ref{tab:throughput-qwen3-8b-full-vocabulary} report the acceptance length and throughput for this additional baseline with Qwen3 8B. We observe that EAGLE-3 reaches 91.0\% of the full vocabulary acceptance length, while \ours{} achieves 95.3\%. Despite the gains in acceptance length from a full vocabulary, the computational overhead leads to lower throughput, corroborating the findings from \citet{zhao-etal-2025-fr}.

\subsection{Kernel Ablation}

To evaluate the impact of our kernel on throughput, we implement a naive indexed LM head operation in PyTorch (Figure~\ref{fig:kernel-diagram}). Table~\ref{tab:throughput-qwen3-8b-kernel-ablation} presents the throughput for Qwen3 8B both with and without our kernel. We observe that using this approach, instead of our kernel, leads to a 2.7\% \emph{decrease} in average throughput compared to EAGLE-3.

\section{Hyperparameters}
\label{app:hyperparameters}

We follow the hyperparameters used by EAGLE-3 \citep{li-etal-2025-eagle3} as closely as possible, deriving values from the paper and software implementation. We report all training hyperparameters in Table~\ref{tab:hyperparameters-training} and all inference hyperparameters in Table~\ref{tab:hyperparameters-inference}.

\section{Datasets}
\label{app:datasets}

We report dataset statistics for our evaluation dataset, Spec-Bench \citep{xia-etal-2024-unlocking}, in Table~\ref{tab:dataset-specbench}. Specifically, we report the number of examples both per task and in aggregate.

\section{Additional Related Work}

Our paper focuses on vocabulary use in neural language models with finite subword vocabularies \citep{abdaoui-etal-2020-load, williams-aletras-2025-vocabulary}. Many speculative decoding approaches rely upon neural language models, typically either using an external draft model \citep{xia-etal-2023-speculative, leviathan-etal-2023-fast, chen-etal-2023-accelerating-large, li-etal-2024-eagle1} or adapting the target model \citep{cai-etal-2024-medusa, zhang-etal-2024-draft, ankner-etal-2024-hydra}. However, some speculative decoding techniques avoid using neural language models for drafting \citep{he-etal-2024-rest, zhao-etal-2024-lookahead, marzollo-etal-2024-sssd}, potentially mitigating the need for vocabulary speculation.

\begin{table}[t]
\small
\centering
\begin{tabular}{lc}
\toprule
Hyperparameter & Value \\
\midrule
Batch Size & 8 \\
Training Steps & 800,000 \\
Warmup Steps & 1.5\% \\
Learning Rate & 5$\times$10\textsuperscript{-5} \\
Adam $\beta_1$ & 0.9 \\
Adam $\beta_2$ & 0.95 \\
Adam $\epsilon$ & 1$\times$10\textsuperscript{-8} \\
Weight Decay & 0 \\
TTT Length & 7 \\
\bottomrule
\end{tabular}
\caption{Hyperparameters used for EAGLE-3 draft model training.}
\label{tab:hyperparameters-training}
\end{table}

\begin{table}[t]
\small
\centering
\begin{tabular}{lc}
\toprule
Hyperparameter & Value \\
\midrule
Decoding Steps & 8 \\
Top-$k$ & 10 \\
Draft Tokens & 60 \\
\bottomrule
\end{tabular}
\caption{Hyperparameters used for inference with EAGLE-3 draft models.}
\label{tab:hyperparameters-inference}
\end{table}

\begin{table}[t!]
\small
\centering
\begin{tabular}{lc}
\toprule
Category & Examples \\
\midrule
Multi-turn Conversation & 80 \\
Machine Translation & 80 \\
Summarization & 80 \\
Question Answering & 80 \\
Mathematical Reasoning & 80 \\
Retrieval-augmented Generation & 80 \\
\midrule
Total & 480 \\
\bottomrule
\end{tabular}
\caption{The number of examples in each category of Spec-Bench \citep{xia-etal-2024-unlocking}.}
\label{tab:dataset-specbench}
\end{table}

\begin{table*}[t]
\small
\centering
\begin{tabular}{l|rrrrrr|r}
\toprule
Method & \multicolumn{1}{c}{Conv.} & \multicolumn{1}{c}{MT} & \multicolumn{1}{c}{Summ.} & \multicolumn{1}{c}{QA} & \multicolumn{1}{c}{Math} & \multicolumn{1}{c}{RAG} & \multicolumn{1}{|c}{Mean} \\
\midrule
Autoregressive Decoding & 1.00 & 1.00 & 1.00 & 1.00 & 1.00 & 1.00 & 1.00 \\
EAGLE-3 (Our Reproduction) & 4.80 & 3.22 & 4.55 & 4.00 & 4.88 & 4.39 & 4.31 \\
\midrule
EAGLE-3 + FR-Spec & 4.56 & 3.41 & 4.77 & 4.04 & 4.87 & 4.32 & 4.33 \\
EAGLE-3 + VocabTrim & 4.83 & 3.24 & 4.64 & 4.12 & 5.08 & 4.39 & 4.38 \\
EAGLE-3 + \ours{} (Ours) & \textbf{4.98} & \textbf{3.43} & \textbf{4.92} & \textbf{4.13} & \textbf{5.14} & \textbf{4.55} & \textbf{4.53} \\
\bottomrule
\end{tabular}
\caption{Acceptance length (tokens) for each decoding method across the Spec-Bench benchmark with OLMo 2 1B. We report the average acceptance length over five seeds. The best result in each category is highlighted in \textbf{bold}.}
\label{tab:acceptance-length-olmo2-1b}
\end{table*}

\begin{table*}[t]
\small
\centering
\begin{tabular}{l|rrrrrr|r}
\toprule
Method & \multicolumn{1}{c}{Conv.} & \multicolumn{1}{c}{MT} & \multicolumn{1}{c}{Summ.} & \multicolumn{1}{c}{QA} & \multicolumn{1}{c}{Math} & \multicolumn{1}{c}{RAG} & \multicolumn{1}{|c}{Mean} \\
\midrule
Autoregressive Decoding & 1.00 & 1.00 & 1.00 & 1.00 & 1.00 & 1.00 & 1.00 \\
EAGLE-3 (Our Reproduction) & 4.76 & 3.58 & 4.39 & 4.19 & 5.09 & 4.44 & 4.41 \\
\midrule
EAGLE-3 + FR-Spec & 4.72 & \textbf{3.87} & 4.54 & 4.19 & 4.93 & 4.52 & 4.46 \\
EAGLE-3 + VocabTrim & 5.03 & 3.74 & 4.44 & 4.24 & 5.21 & 4.51 & 4.53 \\
EAGLE-3 + \ours{} (Ours) & \textbf{5.12} & 3.86 & \textbf{4.75} & \textbf{4.34} & \textbf{5.60} & \textbf{4.75} & \textbf{4.74} \\
\bottomrule
\end{tabular}
\caption{Acceptance length (tokens) for each decoding method across the Spec-Bench benchmark with OLMo 2 7B. We report the average acceptance length over five seeds. The best result in each category is highlighted in \textbf{bold}.}
\label{tab:acceptance-length-olmo2-7b}
\end{table*}

\begin{table*}[t]
\small
\centering
\begin{tabular}{l|rrrrrr|r}
\toprule
Method & \multicolumn{1}{c}{Conv.} & \multicolumn{1}{c}{MT} & \multicolumn{1}{c}{Summ.} & \multicolumn{1}{c}{QA} & \multicolumn{1}{c}{Math} & \multicolumn{1}{c}{RAG} & \multicolumn{1}{|c}{Mean} \\
\midrule
Autoregressive Decoding & 1.00 & 1.00 & 1.00 & 1.00 & 1.00 & 1.00 & 1.00 \\
EAGLE-3 (Our Reproduction) & 5.02 & 3.37 & 4.31 & 4.94 & 5.79 & 4.94 & 4.73 \\
\midrule
EAGLE-3 + FR-Spec & 4.57 & 3.52 & 4.59 & 4.66 & 5.61 & 4.79 & 4.62 \\
EAGLE-3 + VocabTrim & 5.20 & 3.43 & 4.35 & 5.05 & 5.86 & 5.04 & 4.82 \\
EAGLE-3 + \ours{} (Ours) & \textbf{5.26} & \textbf{3.59} & \textbf{4.63} & \textbf{5.09} & \textbf{6.05} & \textbf{5.45} & \textbf{5.01} \\
\bottomrule
\end{tabular}
\caption{Acceptance length (tokens) for each decoding method across the Spec-Bench benchmark with Qwen3 4B. We report the average acceptance length over five seeds. The best result in each category is highlighted in \textbf{bold}.}
\label{tab:acceptance-length-qwen3-4b}
\end{table*}

\clearpage

\begin{table*}[t]
\small
\centering
\begin{tabular}{l|rrrrrr|r}
\toprule
Method & \multicolumn{1}{c}{Conv.} & \multicolumn{1}{c}{MT} & \multicolumn{1}{c}{Summ.} & \multicolumn{1}{c}{QA} & \multicolumn{1}{c}{Math} & \multicolumn{1}{c}{RAG} & \multicolumn{1}{|c}{Mean} \\
\midrule
Autoregressive Decoding & \result{338.3}{3.0} & \result{309.0}{0.2} & \result{330.3}{1.1} & \result{340.6}{0.1} & \result{339.3}{0.4} & \result{332.2}{0.4} & \result{331.6}{0.7} (1.00$\times$) \\
EAGLE-3 (Our Reproduction) & \result{499.5}{1.3} & \result{301.5}{0.7} & \result{457.7}{1.0} & \result{416.0}{0.8} & \result{501.2}{1.1} & \result{445.7}{1.1} & \result{436.9}{1.0} (1.32$\times$) \\
\midrule
EAGLE-3 + FR-Spec & \result{474.0}{0.5} & \result{315.4}{0.7} & \result{477.2}{0.8} & \result{419.1}{0.8} & \result{500.8}{0.5} & \result{438.2}{0.4} & \result{437.4}{0.5} (1.32$\times$) \\
EAGLE-3 + VocabTrim & \result{502.0}{0.9} & \result{302.2}{0.7} & \result{463.7}{1.8} & \result{427.6}{0.9} & \result{519.7}{0.9} & \result{444.7}{1.1} & \result{443.3}{0.8} (1.34$\times$) \\
EAGLE-3 + \ours{} (Ours) & \result{\textbf{519.7}}{0.6} & \result{\textbf{319.0}}{0.5} & \result{\textbf{494.0}}{1.1} & \result{\textbf{430.7}}{0.7} & \result{\textbf{528.2}}{0.9} & \result{\textbf{461.3}}{2.1} & \result{\textbf{458.8}}{0.8} (1.38$\times$) \\
\bottomrule
\end{tabular}
\caption{Throughput (tokens per second) for each decoding method across the Spec-Bench tasks with OLMo 2 1B. We report the average throughput over five seeds, with the standard deviation as subscripts. The best result in each category is highlighted in \textbf{bold}.}
\label{tab:throughput-olmo2-1b}
\end{table*}

\begin{table*}[t]
\small
\centering
\begin{tabular}{l|rrrrrr|r}
\toprule
Method & \multicolumn{1}{c}{Conv.} & \multicolumn{1}{c}{MT} & \multicolumn{1}{c}{Summ.} & \multicolumn{1}{c}{QA} & \multicolumn{1}{c}{Math} & \multicolumn{1}{c}{RAG} & \multicolumn{1}{|c}{Mean} \\
\midrule
Autoregressive Decoding & \result{94.3}{0.0} & \result{88.5}{1.2} & \result{92.1}{0.2} & \result{94.9}{0.4} & \result{95.1}{0.1} & \result{92.6}{0.0} & \result{92.9}{0.2} (1.00$\times$) \\
EAGLE-3 (Our Reproduction) & \result{209.7}{0.1} & \result{140.4}{0.2} & \result{185.9}{0.1} & \result{185.6}{0.1} & \result{224.1}{0.2} & \result{190.3}{0.0} & \result{189.3}{0.1} (2.04$\times$) \\
\midrule
EAGLE-3 + FR-Spec & \result{207.9}{0.1} & \result{150.1}{0.1} & \result{191.9}{0.4} & \result{185.0}{0.9} & \result{217.1}{0.3} & \result{193.2}{0.2} & \result{190.9}{0.1} (2.05$\times$) \\
EAGLE-3 + VocabTrim & \result{221.1}{0.1} & \result{146.0}{0.3} & \result{188.3}{0.2} & \result{187.5}{0.3} & \result{229.3}{0.2} & \result{193.2}{0.1} & \result{194.2}{0.2} (2.09$\times$) \\
EAGLE-3 + \ours{} (Ours) & \result{\textbf{227.4}}{0.1} & \result{\textbf{151.0}}{0.2} & \result{\textbf{202.4}}{0.0} & \result{\textbf{194.0}}{0.1} & \result{\textbf{248.7}}{0.1} & \result{\textbf{204.7}}{0.1} & \result{\textbf{204.7}}{0.0} (2.20$\times$) \\
\bottomrule
\end{tabular}
\caption{Throughput (tokens per second) for each decoding method across the Spec-Bench tasks with OLMo 2 7B. We report the average throughput over five seeds, with the standard deviation as subscripts. The best result in each category is highlighted in \textbf{bold}.}
\label{tab:throughput-olmo2-7b}
\end{table*}

\begin{table*}[t]
\small
\centering
\begin{tabular}{l|rrrrrr|r}
\toprule
Method & \multicolumn{1}{c}{Conv.} & \multicolumn{1}{c}{MT} & \multicolumn{1}{c}{Summ.} & \multicolumn{1}{c}{QA} & \multicolumn{1}{c}{Math} & \multicolumn{1}{c}{RAG} & \multicolumn{1}{|c}{Mean} \\
\midrule
Autoregressive Decoding & \result{160.5}{0.0} & \result{143.4}{0.1} & \result{157.5}{0.0} & \result{161.4}{0.0} & \result{160.6}{0.0} & \result{159.1}{0.0} & \result{157.1}{0.0} (1.00$\times$) \\
EAGLE-3 (Our Reproduction) & \result{357.0}{0.4} & \result{200.2}{0.4} & \result{295.9}{0.2} & \result{353.9}{0.4} & \result{405.8}{0.4} & \result{344.7}{0.8} & \result{326.2}{0.4} (2.08$\times$) \\
\midrule
EAGLE-3 + FR-Spec & \result{326.1}{0.6} & \result{207.5}{0.5} & \result{314.3}{0.6} & \result{334.1}{0.6} & \result{393.9}{1.2} & \result{335.3}{0.6} & \result{318.5}{0.7} (2.03$\times$) \\
EAGLE-3 + VocabTrim & \result{369.4}{0.7} & \result{202.2}{1.2} & \result{298.5}{0.6} & \result{362.0}{0.8} & \result{410.0}{1.1} & \result{351.6}{0.9} & \result{332.3}{0.7} (2.12$\times$) \\
EAGLE-3 + \ours{} (Ours) & \result{\textbf{371.5}}{0.9} & \result{\textbf{207.9}}{3.4} & \result{\textbf{314.8}}{0.7} & \result{\textbf{362.3}}{0.9} & \result{\textbf{421.2}}{1.2} & \result{\textbf{377.4}}{1.1} & \result{\textbf{342.5}}{0.9} (2.18$\times$) \\
\bottomrule
\end{tabular}
\caption{Throughput (tokens per second) for each decoding method across the Spec-Bench tasks with Qwen3 4B. We report the average throughput over five seeds, with the standard deviation as subscripts. The best result in each category is highlighted in \textbf{bold}.}
\label{tab:throughput-qwen3-4b}
\end{table*}

\clearpage

\begin{table*}[t]
\small
\centering
\begin{tabular}{l|rrrrrr|r}
\toprule
Method & \multicolumn{1}{c}{Conv.} & \multicolumn{1}{c}{MT} & \multicolumn{1}{c}{Summ.} & \multicolumn{1}{c}{QA} & \multicolumn{1}{c}{Math} & \multicolumn{1}{c}{RAG} & \multicolumn{1}{|c}{Mean} \\
\midrule
EAGLE-3 (Our Reproduction) & 5.11 & 3.60 & 4.21 & 4.80 & 5.94 & 5.04 & 4.78 \\
EAGLE-3 (Full Vocabulary) & \textbf{5.45} & \textbf{4.01} & \textbf{4.79} & \textbf{5.11} & \textbf{6.41} & \textbf{5.79} & \textbf{5.26} \\
EAGLE-3 + \ours{} (Ours) & 5.30 & 3.82 & 4.49 & 4.92 & 6.28 & 5.26 & 5.01 \\
\bottomrule
\end{tabular}
\caption{Acceptance length (tokens) across the Spec-Bench benchmark with Qwen3 8B, including a full-vocabulary draft model. We report the average acceptance length over five seeds. The best result in each category is highlighted in \textbf{bold}.}
\label{tab:acceptance-length-qwen3-8b-full-vocabulary}
\end{table*}

\begin{table*}[t]
\small
\centering
\begin{tabular}{l|rrrrrr|r}
\toprule
Method & \multicolumn{1}{c}{Conv.} & \multicolumn{1}{c}{MT} & \multicolumn{1}{c}{Summ.} & \multicolumn{1}{c}{QA} & \multicolumn{1}{c}{Math} & \multicolumn{1}{c}{RAG} & \multicolumn{1}{|c}{Mean} \\
\midrule
EAGLE-3 (Our Reproduction) & \result{258.9}{0.1} & \result{159.1}{0.2} & \result{203.9}{0.2} & \result{244.1}{0.2} & \result{297.8}{0.2} & \result{246.9}{0.2} & \result{235.1}{0.2} \\
EAGLE-3 (Full Vocabulary) & \result{219.9}{0.2} & \result{143.0}{0.3} & \result{186.0}{0.2} & \result{206.8}{0.2} & \result{256.4}{0.3} & \result{227.5}{0.2} & \result{206.6}{0.2} \\
EAGLE-3 + \ours{} (Ours) & \result{\textbf{267.6}}{0.2} & \result{\textbf{166.5}}{0.2} & \result{\textbf{216.5}}{0.1} & \result{\textbf{249.4}}{0.2} & \result{\textbf{313.6}}{0.2} & \result{\textbf{257.2}}{0.2} & \result{\textbf{245.2}}{0.2} \\
\bottomrule
\end{tabular}
\caption{Throughput (tokens per second) across the Spec-Bench tasks with Qwen3 8B, including a full-vocabulary draft model. We report the average throughput over five seeds, with the standard deviation as subscripts. The best result in each category is highlighted in \textbf{bold}.}
\label{tab:throughput-qwen3-8b-full-vocabulary}
\end{table*}

\begin{table*}[t]
\small
\centering
\begin{tabular}{l|rrrrrr|r}
\toprule
Method & \multicolumn{1}{c}{Conv.} & \multicolumn{1}{c}{MT} & \multicolumn{1}{c}{Summ.} & \multicolumn{1}{c}{QA} & \multicolumn{1}{c}{Math} & \multicolumn{1}{c}{RAG} & \multicolumn{1}{|c}{Mean} \\
\midrule
EAGLE-3 (Our Reproduction) & \result{258.4}{0.3} & \result{158.5}{0.5} & \result{203.7}{0.3} & \result{243.6}{0.3} & \result{297.2}{0.4} & \result{246.7}{0.3} & \result{234.7}{0.4} \\
EAGLE-3 + \ours{} (Without Kernel) & \result{248.0}{0.2} & \result{154.5}{0.4} & \result{202.0}{0.3} & \result{232.1}{0.2} & \result{287.9}{1.0} & \result{245.2}{0.2} & \result{228.3}{0.3} \\
EAGLE-3 + \ours{} & \result{\textbf{266.9}}{0.2} & \result{\textbf{165.2}}{0.2} & \result{\textbf{215.8}}{0.2} & \result{\textbf{248.6}}{0.1} & \result{\textbf{312.1}}{0.8} & \result{\textbf{256.6}}{0.1} & \result{\textbf{244.2}}{0.2} \\
\bottomrule
\end{tabular}
\caption{Throughput (tokens per second) across the Spec-Bench tasks with Qwen3 8B, including an ablation of our kernel. We report the average throughput over five seeds, with the standard deviation as subscripts. The best result in each category is highlighted in \textbf{bold}.}
\label{tab:throughput-qwen3-8b-kernel-ablation}
\end{table*}

\end{document}